\documentclass[10pt,twocolumn,letterpaper]{article}
\usepackage[accsupp]{axessibility}  
\usepackage{iccv}
\usepackage{times}
\usepackage{epsfig}
\usepackage{graphicx}
\usepackage{amsmath}
\usepackage{amssymb}
\usepackage{caption}
\usepackage{subcaption}
\usepackage{booktabs}
\usepackage{multirow}
\usepackage{color}
\usepackage{xcolor}
\usepackage{colortbl}
\usepackage{comment}
\usepackage{pifont}
\usepackage{array}
\usepackage{url}

\newcolumntype{I}{!{\vrule width 1.2pt}}
\newlength\savedwidth
\newcommand\whline{\noalign{\global\savedwidth\arrayrulewidth
		\global\arrayrulewidth 1.25pt}%
	\hline
	\noalign{\global\arrayrulewidth\savedwidth}}
\newlength\savewidth\newcommand\shline{\noalign{\global\savewidth\arrayrulewidth
  \global\arrayrulewidth 1pt}\hline\noalign{\global\arrayrulewidth\savewidth}}
\newcommand{\tablestyle}[2]{\setlength{\tabcolsep}{#1}\renewcommand{\arraystretch}{#2}\centering\footnotesize}
\renewcommand{\paragraph}[1]{\vspace{1.25mm}\noindent\textbf{#1}}

\newcolumntype{x}[1]{>{\centering\arraybackslash}p{#1pt}}
\newcolumntype{y}[1]{>{\raggedright\arraybackslash}p{#1pt}}
\newcolumntype{z}[1]{>{\raggedleft\arraybackslash}p{#1pt}}

\newcommand{\app}{\raise.17ex\hbox{$\scriptstyle\sim$}}

\definecolor{deemph}{gray}{0.6}

\definecolor{baselinecolor}{gray}{.9}
\newcommand{\baseline}[1]{\cellcolor{baselinecolor}{#1}}

\providecommand{\red}[1]{\textcolor{red}{#1}}
\providecommand{\blue}[1]{\textcolor{blue}{#1}}

\usepackage[pagebackref=true,breaklinks=true,colorlinks,bookmarks=false]{hyperref}

\usepackage[capitalize]{cleveref}
\crefname{section}{Sec.}{Secs.}
\Crefname{section}{Section}{Sections}
\Crefname{table}{Table}{Tables}
\crefname{table}{Tab.}{Tabs.}

\definecolor{mygray}{gray}{.9}
\definecolor{Lavender}{HTML}{E6E6FA}
\definecolor{LightCyan}{HTML}{E1FFFF}

\iccvfinalcopy 

\begin{document}


\title{STEERER: Resolving Scale Variations for Counting and Localization \\ via Selective Inheritance Learning}

\author{
Tao Han$^{1}$, 
Lei Bai$^{1} \thanks{Corresponding author}$, 
Lingbo Liu$^{2}$, 
Wanli Ouyang$^{1}$\\
$^{1}$Shanghai Artificial Intelligence Laboratory, 
$^{2}$The Hong Kong Polytechnic University\\
{\tt\small \{hantao10200, baisanshi,liulingbo918\}@gmail.com, wanli.ouyang@sydney.edu.au}
}

\maketitle

\begin{abstract}
Scale variation is a deep-rooted problem in object counting, which has not been effectively addressed by existing scale-aware algorithms. An important factor is that they typically involve cooperative learning across multi-resolutions, which could be suboptimal for learning the most discriminative features from each scale. In this paper, we propose a novel method termed STEERER (\textbf{S}elec\textbf{T}iv\textbf{E} inh\textbf{ER}itance l\textbf{E}a\textbf{R}ning) that addresses the issue of scale variations in object counting. STEERER selects the most suitable scale for patch objects to boost feature extraction and only inherits discriminative features from lower to higher resolution progressively. The main insights of STEERER are a dedicated Feature Selection and Inheritance Adaptor (FSIA), which selectively forwards scale-customized features at each scale, and a Masked Selection and Inheritance Loss (MSIL) that helps to achieve high-quality density maps across all scales. Our experimental results on nine datasets with counting and localization tasks demonstrate the unprecedented scale generalization ability of STEERER. Code is available at \url{https://github.com/taohan10200/STEERER}.
\end{abstract}

\section{Introduction}
The utilization of computer vision techniques to count objects has garnered significant attention due to its potential in various domains. These domains include but are not limited to, crowd counting for anomaly detection~\cite{li2013anomaly, chaker2017social}, vehicle counting for efficient traffic management~\cite{mundhenk2016large, zhang2017fcn, marsden2018people}, cell counting for accurate disease diagnosis~\cite{eren2022deepcan}, wildlife counting for species protection~\cite{arteta2016counting, laradji2018blobs}, and crop counting for effective production estimation~\cite{lu2017tasselnet}.

\begin{figure}[t]
	\centering
	\includegraphics[width=0.99\linewidth]{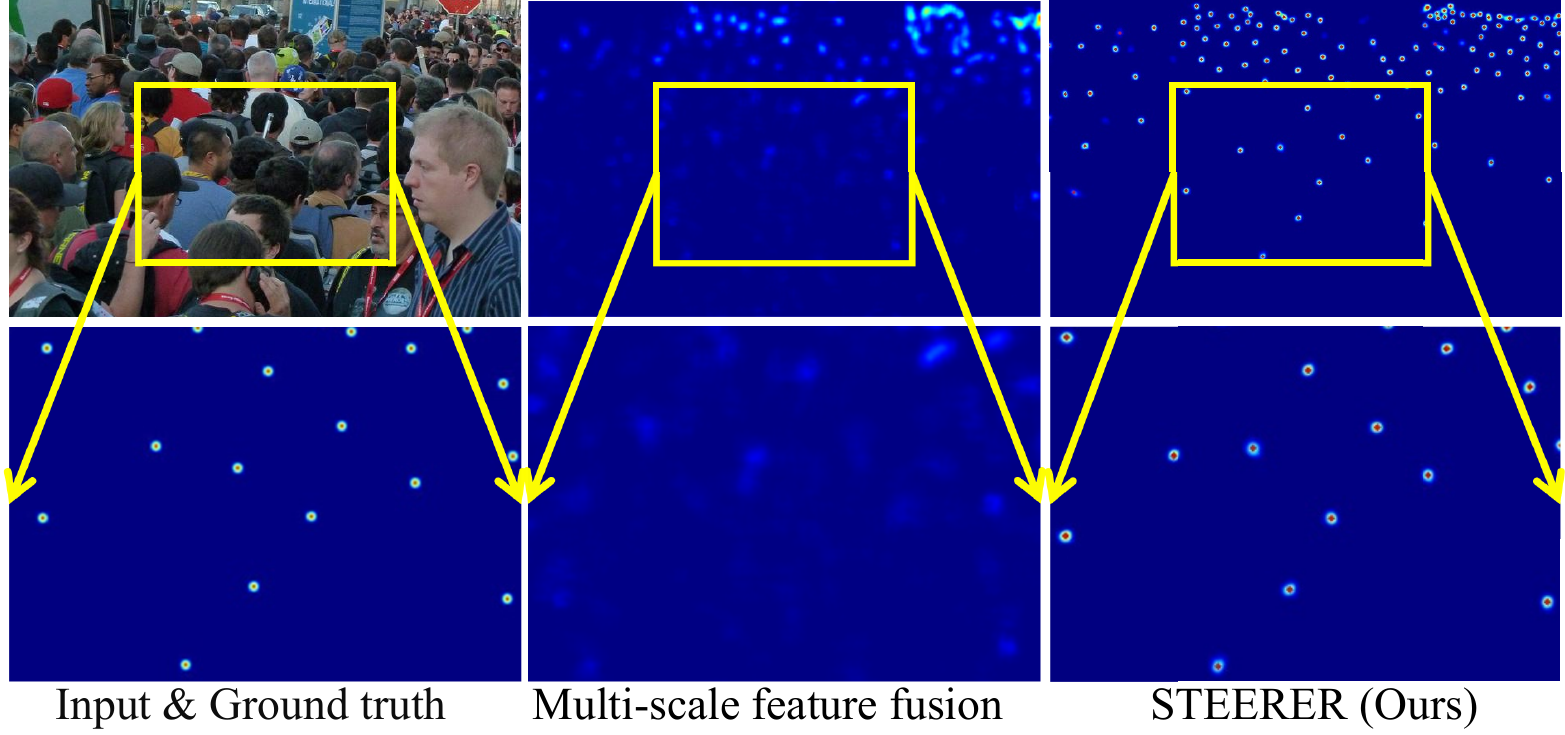}

	\caption{STEERER significantly improves the quality of the predicted density map, especially for large objects, compared with the classic multi-scale future fusion method~\cite{sindagi2019multi}.} 
	\label{fig:multi-scale}

\end{figure}

Numerous pioneers~\cite{Liu_2019_ICCV,sindagi2018survey,sam2020locate,khan2022revisiting,wu2023yunet,ma2020learning,idrees2013multi,zeng2017multi,jiang2023deep} have emphasized that the scale variations encountered in counting tasks present a formidable challenge, motivating the development of various algorithms aimed at mitigating their deleterious impact. These algorithms can be broadly classified into two branches: 1) \textbf{Learn to scale methods}~\cite{liu2019recurrent,Xu_2019_ICCV,sajid2020zoomcount} address scale variations in a two-step manner. Specifically, it involves estimating an appropriate scale factor for a given image/feature patch, followed by resizing the patch for prediction. Most of them require an additional training task (e.g., predicting the object's scale) and the predicted density map is further partitioned into multiple parts,
2) \textbf{Multi-scale fusion methods}~\cite{lin2017feature,liu2018path} have been demonstrated to be effective in handling scale variations in various visual tasks, and their concept is also extensively utilized in object counting. Generally, they focus on two types of fusion, namely feature fusion~\cite{idrees2013multi,zeng2017multi,8354231,sindagi2017generating,cao2018scale,sindagi2019multi,ma2022fusioncount,meng2022hierarchical,li2019crowd,zhao2021msr,9681311} and density map fusion~\cite{du2022redesigning,liu2021exploiting,jiang2020attention,wan2019adaptive,liu2020adaptive,ma2020learning,song2021choose,cheng2019learning}. Despite the recent progress, these studies still face challenges in dealing with scale variations, especially for large objects, as shown in Fig. \ref{fig:multi-scale}. The main reason is that the existing multi-scale methods (\eg FPN~\cite{lin2017feature}) adapt the same loss to optimize all resolutions, which poses a great challenge for each resolution to find which scale range is easy to handle. Also, it results in mutual suppression because counting the same object accurately at all scales is hard.                   

Prior research~\cite{zeng2017multi,liu2021exploiting,song2021choose} has revealed that different resolutions possess varying degrees of scale tolerance in a multi-resolution fusion network. Specifically, a single-resolution feature can accurately represent some objects within a limited scale range, but it may not be effective for other scales, as visualized in feature maps in \cref{fig:framework} \textcolor{red}{red} boxes. We refer to features that can accurately predict object characteristics as scale-customized features; otherwise, they are termed as scale-uncustomized features. If scale-customized features can be separated from their master features before the fusion process, it is possible to preserve the discriminative features throughout the entire fusion process. Hence, our \textbf{first motivation} is to \emph{disentangle each resolution into scale-customized and scale-uncustomized features before each fusion, enabling them to be independently processed. }
To accomplish this, we introduce the Feature Selection and Inheritance Adaptor (FSIA), which comprises three sub-modules with independent parameters. Two adaptation modules separately process the scale-customized and the scale-uncustomized features, respectively, while a soft-mask generator attentively selects features in the middle.

Notably, conventional optimization methods do not ensure that FSIA acquires the desired specialized functions. Therefore, our \textbf{second motivation} is to \emph{enhance FSIA's capabilities through exclusive optimization targets at each resolution}. The first function of these objectives is to implement inheritance learning when transmitting the lower-resolution feature to a higher resolution. This process entails combining the higher-resolution feature with the scale-customized feature disentangled from the lower resolution, which preserves the ability to precisely predict larger objects that are accurately predicted at lower resolutions. Another crucial effect of these objectives is to ensure that each scale is able to effectively capture objects within its proficient scale range, given that the selection of the suitable scale-customized feature at each resolution is imperative for the successful implementation of inheritance learning. 

In conclusion, our ultimate goal is to enhance the capabilities of FSIA through Masked Selection and Inheritance Loss (MSIL), which is composed of two sub-objectives controlled by two mask maps at each scale. To build them, we propose a Patch-Winner Selection Principle that automatically selects a proficient region mask for each scale. The mask is applied to the predicted and ground-truth density maps, enabling the selection of the effective region and filtering out other disturbances. Also, each scale inherits the mask from all resolutions lower than it, allowing for a gradual increase in the objective from low-to-high resolutions, where the incremental objective for a high resolution is the total objective of its neighboring low resolution. The proposed approach is called SeleceTivE inhERitance lEaRning (STEERER), where FSIA and MSIL are combined to maximize the scale generalization ability. STEERER achieves state-of-the-art counting results on several counting tasks and is also extended to achieve SOTA object localization. This paper's primary contributions include:
\begin{itemize}

\item We introduce STEERER, a principled method for object counting that resolves scale variations by cumulatively selecting and inheriting discriminative features from the most suitable scale, thereby enabling the acquisition of scale-customized features from diverse scales for improved prediction accuracy.

\item We propose a Feature Selection and Inheritance Adaptor that explicitly partitions the lower scale feature into its discriminative and undiscriminating components, which facilitates the integration of discriminative representations from lower to higher resolutions and their progressive transmission to the highest resolution.

\item We propose a Masked Selection and Inheritance Loss that utilizes the Patch-Winner Selection Principle to select the optimal scale for each region, thereby maximizing the discriminatory power of the features at the most suitable scales and progressively empowering FSIA with progressive constraints.

\end{itemize}

\section{Related Work}
\subsection{Multi-scale Fusion Methods}
\noindent\textbf{Multi-scale Feature Fusion.} 
This genre aims to address scale variations by leveraging multi-scale features or multi-contextual information~\cite{zhang2016single,sam2017switching,sindagi2017generating,chen2019scale,idrees2013multi,zeng2017multi,8354231,sindagi2017generating,cao2018scale,sindagi2019multi,ma2022fusioncount,meng2022hierarchical,li2019crowd,zhao2021msr,9681311}. They can be further classified into non-attentive and attentive fusion techniques. Non-attentive fusion methods, such as MCNN~\cite{zhang2016single}, employ multi-size filters to generate different receptive fields for scale variations. Similarly, Switch-CNN~\cite{sam2017switching} utilizes a switch classifier to select the optimal column for a given patch. Attentive fusion methods, on the other hand, utilize a visual attention mechanism to fuse multi-scale features. For instance, MBTTBF~\cite{sindagi2019multi} combines multiple shallow and deep features using a self-attention-based fusion module to generate attention maps and weight the four feature maps. Hossain \etal~\cite{hossain2019crowd} propose a scale-aware attention network that automatically focuses on appropriate global and local scales.

\noindent\textbf{Multi-scale Density Fusion.} 
This approach not only adapts multi-scale features but also hierarchically merges multi-scale density maps to improve counting performance~\cite{du2022redesigning,liu2021exploiting,jiang2020attention,Liu_2019_ICCV,wan2019adaptive,liu2020adaptive,ma2020learning,song2021choose,cheng2019learning,kang2018crowd}. Visual attention mechanisms are utilized to regress multiple density maps and extra weight/attention maps~\cite{du2022redesigning,jiang2020attention,song2021choose} during both training and inference stages. For example, DPN-IPSM~\cite{ma2020learning} proposes a weakly supervised probabilistic framework that estimates scale distributions to guide the fusion of multi-scale density maps. On the other hand, Song \etal~\cite{song2021choose} propose an adaptive selection strategy to fuse multiple density maps by selecting region-aware hard pixels through a PRALoss and optimizing them in a fine-grained manner.

\subsection{Learn to Scale Methods} 
These studies typically employ a single-resolution architecture but utilize auxiliary tasks to learn a tailored factor for resizing the feature or image patch to refine the prediction. The final density map is a composite of the patch predictions. For instance, Liu \etal~\cite{liu2019recurrent} propose the Recurrent Attentive Zooming Network, which iteratively identifies regions with high ambiguity and evaluates them in high-resolution space. Similarly, Xu \etal~\cite{Xu_2019_ICCV} introduce a Learning to Scale module that automatically resizes regions with high density for another prediction, improving the quality of the density maps. ZoomCount~\cite{sajid2020zoomcount} categorizes image patches into three groups based on density levels and then uses specially designed patch-makers and crowd regressors for counting.

\textbf{Differences with Previous.}
The majority of scale-aware methods adopt a divide-and-conquer strategy, which is also employed in this study. However, we argue that this does not diminish the novelty of our work, as it is a fundamental thought in numerous computer vision works. Our technical designs significantly diverge from the aforementioned methods and achieve better performance. Typically, Kang \etal~\cite{kang2018crowd} attempts to alter the scale distribution by inputting multi-scale images. In contrast, our method disentangles the most suitable features from multi-resolution representation without multi-resolution inputs, resulting in a distinct general structure. Additionally, some methods \cite{kang2018crowd,Xu_2019_ICCV} require multiple forward passes during inference, while our method only requires a single forward pass.

\begin{figure*}[t]
	\centering
	\includegraphics[width=0.95\textwidth]{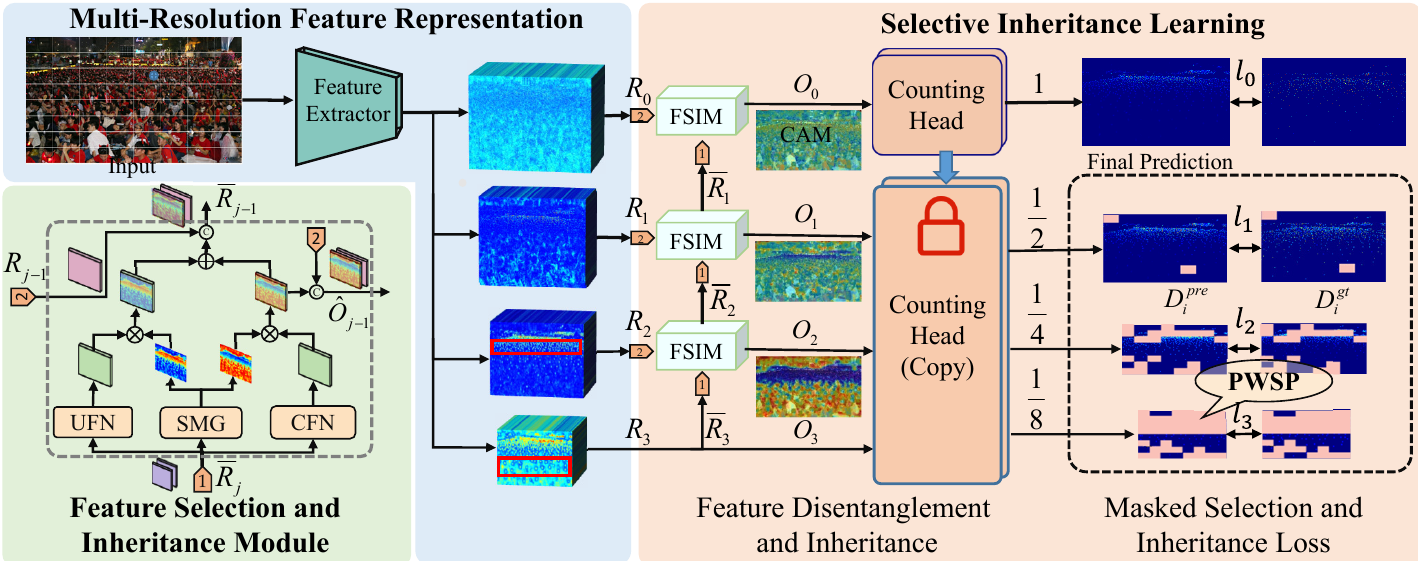}
	\caption{Overview of STEERER. Multi-scale features are fused from the lowest resolution to the highest resolution with the proposed FSIA under the supervision of selective inheritance learning. CAM~\cite{zhou2016learning} figures indicate the proficient regions at each scale. Inference only uses the prediction map from the highest resolution. The masked patches in density maps signify that they are ignored during loss calculation.} 
	\label{fig:framework}
\end{figure*}

\section{Selective Inheritance Learning}

\subsection{Multi-resolution Feature Representation}
Multi-resolution features are commonly adapted in deep learning algorithms to capture multi-scale objects. The feature visualizations in the \textcolor{red}{red} boxes depicted in Fig. \ref{fig:framework} demonstrate that lower-resolution features are highly effective at capturing large objects, while higher-resolution features are only sensitive to small-scale objects. Therefore, building multi-resolution feature representations is critical.
Specifically, let $I \in \mathbb{R}^{3\times H \times W}$ be the input RGB image, and $\Phi_{\theta_{b}}$ denotes the backbone network, where $\theta_{b}$ represents its parameters. We use $\{R_{j}\}_{j=0}^{N}=\Phi_{\theta_{b}}(I)$ to represent $N+1$ multi-feature representations, where the $j$th level's spatial resolution is $(h_j, w_j) = (H/2^{j+2}, W/2^{j+2})$, and $R_N$ represents the lowest resolution feature. The fusion process proceeds from the bottom to the top. This paper primarily utilizes HRNet-W48~\cite{wang2020deep} as the backbone, as its multi-resolution features have nearly identical depths. Additionally, some extended experiments based on the VGG-19 backbone~\cite{simonyan2014very} are conducted to generalize STEERER.

\subsection{Feature Selection and Inheritance Adaptor}
\label{sec:selective_count}
\cref{fig:framework} exemplifies that the feature of large objects in level $P_{3}$ is the most aggregated and most dispersive in level $P_{0}$. Directly upsampling the lowest feature and fusing it to a higher resolution has two drawbacks and we propose FSIA to handle them: 1) the upsampling operations can degrade the scale-customized feature at each resolution, leading to reduced confidence for large objects, as depicted in the middle image of \cref{fig:multi-scale}; 2) the dispersive feature of large objects constitutes noise in higher resolution, and vice versa.

\noindent \textbf{Structure and Functions.} 
The FSIA module, depicted in the lower left corner of \cref{fig:framework}, comprises three components that are learnable, where the scale-Customized feature Forward Network (CFN) and the scale-Uncustomized feature Forward Network (UFN), represented as $C_{\theta{c}}$ and $U_{\theta{u}}$ respectively. They both contain two convolutional layers followed by batch normalization and the Rectified Linear Unit (ReLU) activation function. The Soft-Mask Generator (SMG), $A_{\theta{m}}$, parameterized by $\theta{m}$, is composed of three convolutional layers. The CFN is responsible for combining the upsampled scale-customized features. The UFN, on the other hand, continuously forwards scale-uncustomized features to higher resolutions since they may be potentially beneficial for future disentanglement. If these features are not necessary, the UFN can still suppress them, minimizing their impact. The SMG actively identifies and generates two attention maps for feature disentanglement. Their relationships can be expressed as follows:
\begin{equation}
\small
\begin{array}{cl}
     & O_{j-1} = \text{\textbf{C}}\{R_{j-1}, \mathcal{A}_c \odot C_{\theta{c}}(\bar{R}_{j})\}  \\
     & \bar{R}_{j-1} = \text{\textbf{C}}\{R_{j-1}, \mathcal{A}_u \odot U_{\theta{u}}(\bar{R}_{j})+\mathcal{A}_c \odot C_{\theta{c}}(\bar{R}_{j})\} 
\end{array},
\end{equation}
where $\textbf{C}$ is the feature concatenation, $j=1,2,...,N$ and $\mathcal{A} = \text{Softmax}(A_{\theta{m}}(\bar{R}_j), dim=0), \mathcal{A} \in \mathbb{R}^{2 \times\ h_j \times w_j}$ is a  two-channel attention map. It is split into $\mathcal{A}_c$ and $\mathcal{A}_u$ along the channel dimension. $\odot$ is the Hadamard product. The fusion process begins with the last resolution, where the initial $\bar{R}_3$ is set to $R_3$. 
The input streams for the FSIA are $R_{j-1}$ and $\bar{R}_{j}$. The output streams are $O_{j-1}$ and $\bar{R}_{j-1}$ with the same spatial resolution as $R_{j-1}$. $U_{\theta{u}}$, $C_{\theta{c}}$ and $A_{\theta{m}}$ all have an inner upsampling operation. Notably, in the highest resolution, only the CFN is activated as the features are no longer required to be passed to subsequent scales.

\subsection{Masked Selection and Inheritance Loss}

 FSIA is supposed to be trained toward its functions by imposing some constraints.  In essence, our assumptions for addressing scale variations are 1) \emph{Each resolution can only yield good features within a certain scale range.} and 2) \emph{For the objects belong to the same category, an implicit ground-truth feature distribution $O_j^g$, exists that can be used to infer a ground-truth density map by forwarding it to a well-trained counting head.} 
If $O_j^g$ is available, then the scale-customized feature can be accurately selected from $R^g_j$ and constrained appropriately. However, since the $O_j^g$ is impossible to figure out, we resort to masking the ground-truth density map $D^{gt}_{j}$ for feature selection. To achieve this, we introduce a counting head $E_{\theta_{e}}$, with $\theta_{e}$ being its parameter that is trained solely by the final output branch and kept frozen in other resolutions. By utilizing identical parameters, the feature pattern that can estimate the ground-truth Gaussian kernel is the same at every resolution. For instance, considering the FSIA depicted in Fig. \ref{fig:framework} at the $R_2$ level, we first posit an ideal mask $M^{g}_{2}$ that accurately determines which regions in the $R_2$ level are most appropriate for prediction in comparison to other resolutions. Then the feature selection in $R_2$ level is implemented as, 
 
\begin{equation}
     \ell_{2}^{S} =\mathcal{L}({M^{g}_{2} \odot E_{\theta{e}}(O_2),M^{g}_{2} \odot E_{\theta{e}}(O_2^g) }),
\label{eq:select}
\end{equation}
where $E_{\theta{e}}(O_2^g)$ is the ground truth map. The substitute of the ideal $M_{j}^{g}$ will be elaborated in the PWSP subsection. With such a regulation, ${R_2}$ level will focus on the objects that have the closest distribution with ground truth. Apart from selection ability, $R_2$ level also requires inheriting the scale-customized feature upsampled from $R_3$ level. So another objective in $R_2$ level is to re-evaluate the most suitable regions in $R_3$. That is, we defined $\ell_2^{I}$ on $R_2$ feature level to inherit the scale-customized feature from $R_3$,
\begin{equation}
\small
     \ell_{2}^{I} =\mathcal{L}(\textbf{U}(M^{g}_{3}) \odot E_{\theta{e}}(O_2), \textbf{U}(M^{g}_{3}) \odot E_{\theta{e}}(O^g_2)), 
     \label{eq:inherit}
\end{equation}
where \textbf{U} performs upsampling operation to make the $M^{g}_{3}$ have the same spatial resolution as $E_{\theta{e}}(O_2)$. 

 Hereby, the feature selection and inheritance are supervised by $\ell_{2}^{S}$ and $\ell_{2}^{I}$, respectively. The inheritance loss activates the FSIA, which enables SMG to disentangle the $R_3$ resolution. Simultaneously, it also ensures that the fused feature $O_2$, which combines the $R_2$ level with the scale-customized feature from $R_3$, retains the ability to provide a refined prediction, as achieved by the $R_3$ level. The remaining FSIAs operate similarly, leading to hierarchical selection and inheritance learning. Finally, all objects are aggregated at the highest resolution for the final prediction.

\noindent \textbf{Patch-Winner Selection Principle.}
Another crucial step is about how to determine the ideal mask $M_j^g$ in \cref{eq:select} and \cref{eq:inherit} for each level. Previous studies~\cite{Xu_2019_ICCV,sajid2020zoomcount,ma2020learning} employ scale labels to train scale-aware counting methods, where the scale label is generated in accordance with the geometric distribution~\cite{ma2020learning,ma2019bayesian} or density level~\cite{jiang2020attention,sajid2020zoomcount,sindagi2017generating,Xu_2019_ICCV}. However, they only approximately hold when the objects are evenly distributed. 
Thus, instead of generating $M_j^g$ by manually setting hyperparameters, we propose to allow the network to determine the scale division on its own. That is, each resolution automatically selects the most appropriate region as a mask. Our counting method is based on the popular density map estimation approach~\cite{lempitsky2010learning,pham2015count,xu2016crowd,wang2021neuron,han2020focus}, which transfers point annotations $\mathbf{P^{gt}} = \{(x_i,y_i)\}_{i=0}^M$ into a 2D Gaussian-kernel based density map as ground-truth. As depicted in Fig.~\ref{fig:framework}, the $j$th resolution has a ground-truth density map denoted as $D_{j}^{gt} \in \mathbb{R}^{h_j \times w_j}$, where $\mathbf{P^{gt}}$ is divided by a down-sampling factor $2^{j}$ in each scale.

\begin{figure}[t]
	\centering
		\includegraphics[width=0.98\linewidth]{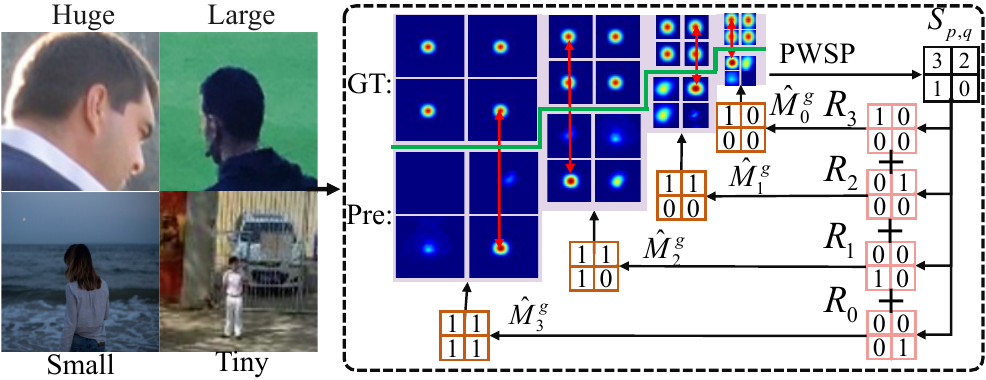}
		\caption{An example to illustrate the PWSP and mask inheriting approaches.}
		\label{fig:PWST}
\end{figure}

\begin{table*}[ht]
	\centering
	\small
	\caption{Leaderboard of NWPU-Crowd counting (\emph{test set}). The best and second-best are shown in \red{red} and \blue{blue}, respectively.}
	\label{tab:nwpu_counting}
	\def\arraystretch{1.}
	\resizebox{\linewidth}{!}{
		\begin{tabular}{@{}l|c|ccc|c|ccccc|c|ccc@{}}
			\toprule
			\multirow{2}{*}{Method} & \multirow{2}{*}{Venue} & & {Overall} & \multicolumn{6}{c|}{Scene Level (MAE)} & \multicolumn{4}{c}{Luminance (MAE)}
			\\\cmidrule{3-15}
			& & MAE & MSE & NAE & Avg. & S0 & S1 & S2 & S3 & S4 & Avg. & L0 & L1 & L2 \\
			\midrule
			MCNN~\cite{zhang2016single} & CVPR16  & \cellcolor{black!10}232.5 & \cellcolor{black!10}714.6 & \cellcolor{black!10}1.063 & 1171.9 & 356.0 & 72.1 & 103.5 & 509.5 & 4818.2 & 220.9 & 472.9 & 230.1 & 181.6 \\
			CSRNet~\cite{li2018csrnet}  & CVPR18  & \cellcolor{black!10}121.3 & \cellcolor{black!10}387.8 & \cellcolor{black!10}0.604 & 522.7 & 176.0 & 35.8 & 59.8  & 285.8 & 2055.8 & 112.0 & 232.4 & 121.0 & 95.5 \\
			CAN~\cite{liu2019context}   & CVPR19  & \cellcolor{black!10}106.3 & \cellcolor{black!10}386.5 & \cellcolor{black!10}0.295 &612.2 &82.6  & 14.7 & 46.6  & 269.7 & 2647.0 & 102.1 & 222.1 & 104.9 & 82.3 \\
			BL~\cite{ma2019bayesian}    & ICCV19  & \cellcolor{black!10}105.4 & \cellcolor{black!10}454.2 & \cellcolor{black!10}0.203 & 750.5 & 66.5 & 8.7 & 41.2 & 249.9 & 3386.4 & 115.8 & 293.4 & 102.7 & 68.0 \\
			SFCN+~\cite{wang2020nwpu}        & PAMI20 & \cellcolor{black!10}105.7 & \cellcolor{black!10}424.1 & \cellcolor{black!10}0.254 & 712.7 & 54.2 & 14.8 & 44.4 & 249.6 & 3200.5 & 106.8 & 245.9 & 103.4 & 78.8 \\
			DM-Count~\cite{wang2020distribution} & NeurIPS20 & \cellcolor{black!10}88.4 & \cellcolor{black!10}388.6 & \cellcolor{black!10}0.169 & 498.0 & 146.7 & 7.6 & 31.2 & 228.7 & 2075.8 & 88.0 & 203.6 & 88.1 & 61.2 \\
			UOT~\cite{ma2021learning}   & AAAI21 &  \cellcolor{black!10}87.8 & \cellcolor{black!10}387.5 & \cellcolor{black!10}0.185 & 566.5 & 80.7 & 7.9 & 36.3 & 212.0 & 2495.4 &95.2 &240.3 & 86.4 & 54.9 \\
			GL~\cite{wan2021generalized}     & CVPR21 & \cellcolor{black!10}79.3 & \cellcolor{black!10}346.1 & \cellcolor{black!10}0.180 & 508.5 & 92.4 & 8.2 & 35.4 & 179.2 & 2228.3 & 85.6 & 216.6 & 78.6 & 48.0 \\
			D2CNet~\cite{cheng2021decoupled}       & IEEE-TIP21& \cellcolor{black!10}85.5 & \cellcolor{black!10}361.5 & \cellcolor{black!10}0.221 & 539.9 & 52.4 & 10.8 &36.2 &212.2 &2387.8 & 82.0 & 177.0 & 83.9 & 68.2\\

			P2PNet~\cite{song2021rethinking} & ICCV21 & \cellcolor{black!10}72.6  & \cellcolor{black!10}331.6 & \cellcolor{black!10}0.192 & 510.0 & \red{\textbf{34.7}} & 11.3 & 31.5 & 161.0 & 2311.6 & 80.6 & 203.8 & 69.6 & 50.1 \\
			MAN~\cite{lin2022boosting} & CVPR22 & \cellcolor{black!10}76.5& \cellcolor{black!10}323.0 & \cellcolor{black!10}0.170  &464.6 &\blue{\textbf{43.3}} &8.5& 35.3&190.9 &2044.9 &  76.4& 180.1&77.1& 49.4\\
		    Chfl~\cite{shu2022crowd}& CVPR22 &\cellcolor{black!10} 76.8&\cellcolor{black!10} 343.0 & \cellcolor{black!10}0.171     &470.0 &56.7 &8.4 &32.1 & 195.1 &2058.0 & 85.2& 217.7&74.5 &49.6 \\
		\midrule
			 STEERER-VGG19 &  -   &\cellcolor{black!10} \blue{\textbf{68.3}}  & \cellcolor{black!10}\blue{\textbf{318.4}} &\cellcolor{black!10} \blue{\textbf{0.165}}  &\blue{\textbf{438.2}} &  61.9 & \blue{\textbf{8.3}} & \blue{\textbf{29.5}} &\blue{\textbf{159.0}} & \blue{\textbf{1932.1}} & \blue{\textbf{71.0}}&  \blue{\textbf{171.3}}& \blue{\textbf{67.4}}& \blue{\textbf{46.2}} \\
			
			STEERER-HRNet  &  -  &\cellcolor{black!10} \red{\textbf{63.7}}  & \cellcolor{black!10}\red{\textbf{309.8}} &\cellcolor{black!10} \red{\textbf{0.133}}  &\red{\textbf{410.6}} &  48.2 & \red{\textbf{6.0}} & \red{\textbf{25.8}} &\red{\textbf{158.3}} & \blue{\textbf{1814.5}} & \red{\textbf{65.1}}&  \red{\textbf{155.7}}& \red{\textbf{63.3}}& \red{\textbf{42.5}} \\
			\bottomrule
		\end{tabular}
	}
\end{table*}

In detail, we propose a Patch-Winner Selection Principle (PWSP) to let each resolution select its capable region, whose thought is to find which resolution has the minimum cost for a given patch. During the training phase, each resolution outputs its predicted density map $D_j^{pre}$, where it is with the same spatial resolution as $D^{gt}_{j}$. As shown in \cref{fig:PWST}, each pair of $D^{gt}_{j}$ and $D^{pre}_{j}$ are divided into $P \times Q$ regions, where $(P, Q)=(\frac{h_j}{h^{'}_j}, \frac{w_j}{w^{'}_j})$, $(h^{'}_j, w^{'}_j)=(\frac{h_0}{2^j}, \frac{w_0}{2^j})$ is the patch size, and we empirically set the patch size $(h_0,w_0)=(256,256)$. (Note that the image will be padded to be divisible by this patch size during inference.) PWSP finally decides the best resolution for a given patch by comparing a re-weighted loss among four resolutions, which is defined as \cref{eq:selected_metric}, 
\begin{equation}
\small
S_{p,q}=\arg \min_{j} \left(\frac{\left\|y_{p,q}^j-\widehat{y}_{p,q}^j\right\|_2^2}{h_j^p \times w_j^p}+ \frac{\left\|y_{p,q}^j-\widehat{y}_{p,q}^j\right\|_2^2}{\left\|y_{p,q}^j\right\|_1+e}\right),
\label{eq:selected_metric}
\end{equation}
where the first item is the Averaged Mean Square Error (AMSE) between ground-truth density patch $y_{p,q}^j$ and predicted density patch $\widehat{y}_{p,q}^j$, and the second item is the Instance Mean Square Error (IMSE). AMSE inclines to measure the overall difference, and it still works when a patch has no object, whereas IMSE gives emphasis on the foreground. $e$ is a very small number to prevent pointless division.   

\noindent \textbf{Total Optimization Loss.}
\label{sec:mrhc}
PWSP dynamically gets the region selection label $S_{p,q}$. \cref{fig:PWST} shows that $S_{p,q}$ are transferred to mask $M_j^g$ with a one-hot encoding operation and an upsampling operation, namely, $\{\bar{M}_{j}^{g}\}_{j=0}^{N} = \texttt{scatter}(S_{p,q})$, where $\bar{M}_j^g(p,q) \in \{0,1\} $ and $\sum^N_{j=0}{\bar{M}_j^g(p,g)}=1$. Hereby, the final scale selection and inherit mask for each resolution is obtained by \cref{eq:mask_reduced}, 
\begin{equation}
\hat{M}_{j}^g = \sum_{j=0}^{N-j} \bar{M}_{j}^{g},
\label{eq:mask_reduced}
\end{equation}
where $N$ is resolution number and  $\hat{M}_{j}^{g} \in \mathbb{R}^{P \times Q}$, Finally, $\hat{M}_j^g$ is interpolated to be $M_j^g$ in \cref{eq:select} and \cref{eq:inherit}, namely $M_j^g = \textbf{U}(\hat{M}_j^g)$, which has the same spatial dimension with $D^{gt}_j$ and $D^{pre}_{j}$. The $\ell_j^S$ and $\ell_j^I$ in \cref{eq:select} and \cref{eq:inherit} can be summed to a single loss $\ell_j$ by adding their masked weight. The ultimate optimization objective for STEERER is: 
\begin{equation}
\centering
l = \sum_{j=1}^{N} \alpha_j \ell_{j}= \sum_{j=1}^{N} \alpha_j  \mathcal{L}(M_{j}^{g} \odot D^{gt}_j, M_{j}^{g} \odot D^{pre}_{j} ),
\label{eq:final_objective}
\end{equation}
where $\alpha_j$ is the weight at each resolution, and we empirically set it as $\alpha_j = 1/2^{j}$. $\mathcal{L}$ is the Euclidean distance.

\begin{table}[t]
	\centering
	\caption{Crowd counting performance on SHHA, SHHB, UCF-QNRF, and JHU-CROWD++ datasets.}
	\label{tab:counting_results}

	\resizebox{1.0\linewidth}{!}{
	\tablestyle{1.5pt}{1.0}
	\begin{tabular}{@{}l|cc|cc|cc|cc@{}}
		\toprule
		\multirow{2}{*}{Method}  & \multicolumn{2}{c|}{SHHA} & \multicolumn{2}{c|}{SHHB} & \multicolumn{2}{c|}{UCF-QNRF} & \multicolumn{2}{c}{JHU-CROWD++}\\
		\cmidrule{2-9}  & MAE & MSE & MAE & MSE & MAE & MSE &MAE & MSE  \\
		\midrule
		CAN~\cite{liu2019context}            & 62.3 & 100.0 & 7.8  & 12.2 & 107.0 & 183.0 & 100.1 &314.0\\
		SFCN~\cite{wang2019learning}      & 64.8 & 107.5 & 7.6  & 13.0 & 102.0 & 171.4 & 77.5 &297.6\\
		S-DCNet~\cite{Xiong_2019_ICCV}         & 58.3 & 95.0  & 6.7  & 10.7 & 104.4 & 176.1&-&- \\
		BL~\cite{ma2019bayesian}             & 62.8 & 101.8 & 7.7  & 12.7 & 88.7 & 154.8  & 75.0 &299.9\\
		ASNet~\cite{jiang2020attention}       & 57.8 & 90.1  & -  &-   & 91.6 & 159.7&-&- \\
		AMRNet~\cite{liu2020adaptive}          & 61.5 & 98.3  & 7.0 & 11.0 & 86.6 & 152.2 &-&-\\
		DM-Count~\cite{wang2020distribution}   & 59.7 & 95.7 & 7.4 & 11.8 & 85.6 & 148.3 &-&-\\
		GL~\cite{wan2021generalized}                & 61.3 & 95.4 & 7.3 & 11.7 & 84.3 & 147.5&-&- \\
		D2CNet~\cite{cheng2021decoupled}        & 57.2 & 93.0 & 6.3 & 10.7 & 81.7 & 137.9 &73.7& 292.5\\
		P2PNet~\cite{song2021rethinking}       & \red{\textbf{52.7}} & \red{\textbf{85.1}} & 6.3 & 9.9 & 85.3 & 154.5 & - &-\\
		SDA+DM~\cite{ma2021towards}            & 55.0 & 92.7 & - & - & 80.7 & 146.3 & 59.3 &248.9\\
    
		MAN~\cite{lin2022boosting} & 56.8& 90.3 & -& - & 77.3& \blue{\textbf{131.5}} & \red{\textbf{53.4}} &\red{\textbf{209.9}}\\
		Chfl~\cite{shu2022crowd}   & 57.5& 94.3 & 6.9& 11.0 & 80.3& 137.6 & 57.0 &235.7\\
	    RSI-ResNet50~\cite{cheng2022rethinking} &{54.8} & {89.1} &\blue{\textbf{6.2}}  &\blue{\textbf{9.9}} & 81.6& 153.7 &58.2&245.1\\	
		\midrule

		STEERER-VGG19                 & 55.6& 87.3 & 6.8 & 10.7 & \blue{\textbf{76.7}} & 135.1 & 55.4 & \blue{\textbf{221.4}} \\

		STEERER-HRNet      & \blue{\textbf{54.5}}& \blue{\textbf{86.9}} & \red{\textbf{5.8}} & \red{\textbf{8.5}} & \red{\textbf{74.3}} &\red{\textbf{128.3}} & \blue{\textbf{54.3}}&238.3\\

		\bottomrule
	\end{tabular}}
\end{table}

\begin{figure*}[htbp]
	\centering
		\includegraphics[width=0.98\linewidth]{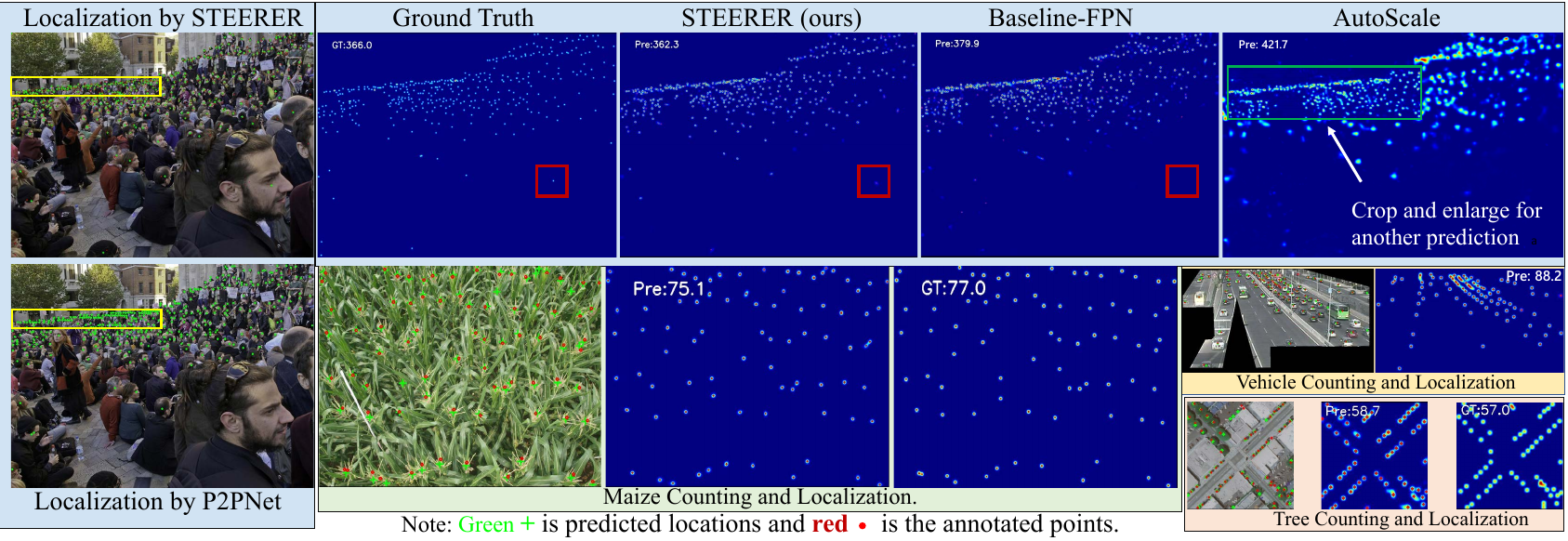}
		\caption{Visualization results of STEERER in counting and localization tasks. GT: ground truth count, Pre: predicted count.}
		\label{fig:vis_all}
\end{figure*}

\section{Experimentation}

\subsection{Experimental Settings}
\noindent\textbf{Datasets.}
STEERER is comprehensively evaluated with object counting and localization experiments on nine datasets: NWPU-Crowd~\cite{wang2020nwpu}, SHHA~\cite{zhang2016single}, SHHB~\cite{zhang2016single}, UCF-QNRF~\cite{idrees2018composition}, FDST~\cite{fang2019locality}, JHU-CROWD++~\cite{sindagi2020jhu}, MTC~\cite{lu2017tasselnet}, URBAN\_TREE~\cite{ventura2022individual}, TRANCOS~ \cite{guerrero2015extremely}. 


\noindent\textbf{Implementation Details.} 
We employ the Adam optimizer~\cite{kingma2014adam}, with the learning rate gradually increasing from $0$ to $10^{-5}$ during the initial 10 epochs using a linear warm-up strategy, followed by a Cosine decay strategy. Our approach is implemented using the PyTorch framework~\cite{paszke2019pytorch} on a single NVIDIA Tesla A100 GPU.

\noindent\textbf{Evaluation Metrics.} For object counting, Mean Absolute Error (MAE)
and Mean Square Error (MSE) are used as evaluation measurements. For localization, we follow the NWPU-Crowd~\cite{wang2020nwpu} localization challenge to calculate instance-level Precision, Recall, and F1-measure to evaluate models. All criteria are defined in the supplementary.

\subsection{Object Counting Results}
\noindent \textbf{Crowd Counting.} NWPU-Crowd benchmark is a challenging dataset with the largest scale variations, which provides a fair platform for evaluation. \cref{tab:nwpu_counting} compares STEERER with the CNN-based SoTAs. STEERER, based on VGG-19 and the HRNet backbone, both outperform other approaches on most of the counting metrics. On the whole, STEERER improves the MAE across all the sub-categories except for the $S0$ level and overall dataset consistently. In detail, STEERER reduces the MAE to \red{63.7} on the NWPU-Crowd test set, which is reduced by $\red{12.3\%}$ compared with the second-place P2PNet~\cite{song2021rethinking}. Note that we currently rank No.1 on the public leaderboard of the NWPU-Crowd benchmark. On other crowd counting datasets, such as  SHHA, SHHB, UCF-QNRF, and JHU-CROWD++, STEERER also achieves excellent performance compared with SoTAs.

\begin{table*}[htbp]
	\centering
	\small
	\caption{The leaderboard of NWPU-Crowd localization (\emph{test set}).}

	\resizebox{1.\linewidth}{!}{
	\tablestyle{12pt}{1.0}
	\begin{tabular}{@{}l|c|ccc|c|r@{}}
		\whline
		\multirow{2}{*}{Method}	&\multirow{2}{*}{Backbone}  &\multicolumn{3}{c|}{Overall ($\sigma_l$)}  &\multicolumn{2}{c}{Box Level (only Rec under $\sigma_l$) (\%)}  \\
		\cline{3-7}
		& & \textbf{F1-m}(\%) & Pre(\%) & Rec(\%)  & Avg. &\text{Head Area: }$A0 \sim A5$\\
		\midrule
		VGG+GPR~\cite{gao2019c,gao2019domain}  &VGG-16 & 52.5&55.8&49.6  & 37.4 & 3.1/27.2/49.1/68.7/49.8/26.3 \\

		RAZ\_Loc~\cite{liu2019recurrent} &VGG-16 &59.8&66.6&54.3 &42.4 & 5.1/28.2/52.0/79.7/64.3/25.1   \\
        Crowd-SDNet~\cite{wang2021self} &ResNet-50 &63.7&65.1&62.4 &55.1& 7.3/43.7/62.4/75.7/71.2/70.2\\
        
        TopoCount~\cite{abousamra2020localization} &VGG-16 & 69.2&68.3&70.1  & \textbf{63.3} & 5.7/39.1/72.2/\textbf{85.7}/\textbf{87.3}/\textbf{89.7} \\

		FIDTM~\cite{liang2022focal} &HRNet & 75.5&79.8&71.7   & 47.5 & \textbf{22.8}/\textbf{66.8}/76.0/71.9/37.4/10.2 \\

          IIM~\cite{gao2020learning}&HRNet  &76.0&\textbf{82.9}&70.2  &49.1 & 11.7/45.3/73.4/83.0/64.5/16.7 \\
LDC-Net~\cite{han2021ldc}&HRNet &76.3&78.5&74.3& 56.6&14.8/53.0/\textbf{77.0}/85.2/70.8/39.0 \\
		\hline
		STEERER  &HRNet  &\textbf{77.0}&81.4&\textbf{73.0} &61.3 & 12.0/46.0/73.2/85.5/86.7/64.3   \\
		\bottomrule
	\end{tabular}
	}
	\label{tab:nwpu_loc}
\end{table*}

\begin{table*}[htbp]
	\centering
	\small
	\caption{Comparison of the crowd localization performance on other crowd datasets.}

    \resizebox{.9\linewidth}{!}{
        \tablestyle{7pt}{0.95}
	  \begin{tabular}{@{}lc|ccc|ccc|ccc|ccr@{}}
		\toprule
		\multirow{2}{*}{Method} & \multirow{2}{*}{Backbone} &\multicolumn{3}{c|}{ShanghaiTech Part A} &\multicolumn{3}{c|}{ShanghaiTech Part B} &\multicolumn{3}{c|}{UCF-QNRF} &\multicolumn{3}{c}{FDST}\\
		\cline{3-14}
		&& \textbf{F1-m} & Pre. &Rec. &\textbf{F1-m} & Pre. &Rec. & \textbf{F1-m} & Pre. &Rec. &\textbf{ F1-m} & Pre. &Rec.\\
	   \midrule
		TinyFaces~\cite{hu2017finding} &ResNet-101&57.3 &43.1 &\textbf{85.5} &71.1 &64.7 &79.0 &49.4 &36.3 &\textbf{77.3} &85.8 &86.1 &85.4  \\

		RAZ\_Loc~\cite{liu2019recurrent} &VGG-16 &69.2 &61.3 &79.5 &68.0  &60.0 &78.3 &53.3 &59.4  &48.3 &83.7 &74.4&95.8  \\
		
		LSC-CNN~\cite{sam2020locate} &VGG-16 &68.0&69.6 &66.5 &71.2 &71.7 &70.6 &58.2 &58.6 &57.7 &- &- &-  \\
	
		IIM~\cite{gao2020learning} &HRNet &73.3 &76.3 &70.5 &83.8 &\textbf{89.8} &78.6 &71.8 &73.7&70.1 &95.4 &95.4&95.3  \\
   
		\midrule
		STEERER &HRNet &\textbf{79.8} &\textbf{80.0} &79.4 &\textbf{87.0} &89.4 &\textbf{84.8} &\textbf{75.5} &\textbf{78.6} &72.7 &\textbf{96.8} &\textbf{96.6} &\textbf{97.0}   \\
		\bottomrule
	\end{tabular}}
	\label{tab:other_loc}
\end{table*}

\noindent \textbf{Plant and Vehicle Counting.}
 \cref{tab:vehicleandplant} shows STEERER works well when directly applying it to evaluate the vehicle (TRANCOS~\cite{guerrero2015extremely}) and Maize counting (MTC~\cite{lu2017tasselnet}). For vehicle counting, we further lower the estimated MAE. For plant counting, our model surpasses other SoTAs, whose MAE and MSE are decreased by $\red{12.9\%}$ and $\red{14.0\%}$, respectively, which shows significant improvements.

\begin{table}
\centering
 \caption{Results on TRANCOS~\cite{guerrero2015extremely} and MTC~\cite{lu2017tasselnet}.}
\resizebox{1.0\linewidth}{!}{
    \tablestyle{14pt}{0.95}
    \begin{tabular}{@{}lcccc@{}}
        \toprule 
        \multirow{2}{*}{\text { Methods }} & \multicolumn{2}{c}{\text { TRANCOS }} & \multicolumn{2}{c@{}}{\text {MTC}} \\
        & \text {MAE} & \text { MSE } & \text { MAE } & \text { MSE } \\
        \midrule
        \text{FCN-HA}~\cite{zhang2017fcn}   & 4.2 &- &- &-  \\
        \text{TasselNetv2}~\cite{xiong2019tasselnetv2} & - & - & 5.4 & 8.8 \\
        \text{S-DCNet}~\cite{Xiong_2019_ICCV} & 2.9 & - & 5.6 & 9.1 \\
        \text{CSRNet }~\cite{li2018csrnet} & 3.6 & - & 9.4 & 14.4 \\
        \text{RSI-ResNet}~\cite{cheng2022rethinking} & 2.1 & \textbf{2.6} &3.1 & 4.3 \\
        \midrule
        \text {STEERER} &\textbf{1.8} & 3.0 & \textbf{2.7} & \textbf{3.7} \\
        \bottomrule
        \end{tabular}
        }
   
    \label{tab:vehicleandplant}
\end{table}

\subsection{Object Localization Results}
\label{Sec:counting_exp}
\noindent \textbf{Crowd Localization.} 
STEERER, benefiting from its high-resolution and elaborated density map, can use a heuristic localization algorithm to locate the object center. Specifically, we follow DRNet~\cite{han2022drvic} to extract the local maximum as a head position from the predicted density map. \cref{tab:nwpu_loc} tabulates the overall performances (Localization: F1-m, Pre. and Rec.) and per-class Rec. at the box level. By comparing the primary key (Overall F1-m) for ranking, our method achieves first place on the F1-m 77.0\%  compared with other crowd localization methods. Notably, Per-class Rec. on Box Level shows the sensitivity of the model for scale variations. The box-level results show our method achieves a balanced performance on all scales (A0$\sim$A5), which further verifies the effectiveness of STEERER for addressing scale variations. 
\cref{tab:other_loc} lists the localization results on other datasets. Here, we adopt the same evaluation protocol with IIM~\cite{gao2020learning}. \cref{tab:other_loc} demonstrates that STEERER refreshes the F1-measure on all datasets. In general, our localization results precede other SoTA methods. Take F1-m as an example. The relative performance of our method is increased by an average of $\red{5.2\%}$ on these four datasets. 

\noindent\textbf{Urban Tree Localization.} 
URBAN$\_$TREEC~\cite{ventura2022individual} collects trees in urban environments with aerial imagery. As presented in~\cref{tab:URBAN_TREE}, ITC~\cite{ventura2022individual} takes multi-spectral imagery as input and outputs a confidence map indicating the locations of trees. The individual tree locations are found by local peak finding. Similarly, the predicted density map in our framework can also be used to locate the urban trees in this way. Our method achieves a significant improvement over the RGB format in ITC~\cite{ventura2022individual}, with $75.0\%$ of the detected trees matching actual trees, and $74.9\%$ of the trees area being detected, despite using fewer input channels.

\begin{table}
\centering
\caption{Localization Results on URBAN\_TREE~\cite{ventura2022individual}}
\resizebox{1.\linewidth}{!}{
\tablestyle{15pt}{0.91}
\begin{tabular}{@{}lcccc@{}}
\toprule 
Methods & Input & F1-m & Pre. &Rec. \\
\midrule
\multirow{2}{*}{ITC~\cite{ventura2022individual}} & RGB & 71.1& 69.5 & 72.8\\
                                                  & RGBNV & 73.5& 73.6 & 73.3\\
\midrule
\text {STEERER} &RGB & \textbf{75.0} & \textbf{75.0} & \textbf{74.9} \\
\bottomrule
\end{tabular}
}
\label{tab:URBAN_TREE}
\end{table}

\subsection{Ablation Study}
The ablations quantitatively compare the contribution of each component. Here, we set our two baselines as, 1)
BL1: Upsample and concatenate them to the first branch. 2)BL2: Using FPN~\cite{lin2017feature} to make bottom-to-up fusion.

\noindent \textbf{Effectiveness of FSIA.} We here compare the proposed FSIA with the traditional fusion methods, namely BL1 and BL2. \cref{tab:fusion methods} shows that the proposed fusion method is able to catch more objects with less computation. Furthermore, FSIA has the utmost efficiency when three components of FSIA are combined into a unit.    

\noindent \textbf{Effectiveness of Hierarchy Constrain.} \cref{tab:loss_conbination} shows that imposing the scale selection and inheritance loss on each resolution can improve performance step by step. The best performance requires optimizing on all resolutions.  

\noindent \textbf{Effectiveness of MSIL.} \cref{tab:mask_ablation} explores the necessity of both selection and inheritance at each scale. If we simply force each scale to ambiguously learn all objects (\cref{tab:mask_ablation} Row 1), FISM does not achieve the function that we want. If we only select the most suitable patch for each scale to learn, it would be crashed (Row 2). So we conclude that selection and inheritance are indispensable in our framework.

\noindent \textbf{Effectiveness of Head Sharing.} The shared counting head ensures that each scale selects its proficient regions equitably. Otherwise, it does not work, as evident from the head-dependent results presented in \cref{tab:mask_ablation}. 
More ablation experiments and baseline results are provided in the supplementary.

\begin{table*}[t]

\centering
\subfloat[
\textbf{Fusion methods}. FDM is more accurate and faster than traditional fusion methods.
\label{tab:fusion methods}
]{
\centering
\begin{minipage}{0.29\linewidth}{\begin{center}
\tablestyle{4pt}{1.05}
\begin{tabular}{@{}lx{24}x{24}x{24}}
Fusion method & MAE & MSE & FLOPs\\
\shline
BL1-Concat                    & 60.4   & 98.5  & 1.21$\times$\\
BL2-FPN~\cite{lin2017feature} & 60.4   & 96.3  & 1.16$\times$\\
CFN                       & 55.3   & 89.7  & \textbf{~0.99$\times$}\\
UFN+CFN                   & 55.1   & 90.5  & \textbf{~0.99$\times$}\\
UFN+CFN+SMG  & \baseline{\textbf{54.6}} & \baseline{\textbf{86.9}} & \baseline{1.00$\times$}
\end{tabular}
\end{center}}\end{minipage}
}
\hspace{2em}
\subfloat[
\textbf{Resolution number}. Increasing constrain gradually can improve counting accuracy.
\label{tab:loss_conbination}
]{
\begin{minipage}{0.29\linewidth}{\begin{center}
\tablestyle{4pt}{1.05}
\begin{tabular}{@{}x{60}x{28}x{28}}
$N$  & MAE & MSE \\
\shline
$\ell_0$ & 58.4 & 92.0 \\
$\ell_0$+$\ell_1$ & 56.5 & 92.8 \\
$\ell_0$+$\ell_1$+$\ell_2$ & 55.8 & 88.1 \\
$\ell_0$+$\ell_1$+$\ell_2$+$\ell_3$ & \baseline{\textbf{54.6}} & \baseline{\textbf{86.9}} \\
\multicolumn{3}{c}{~}\\
\end{tabular}
\end{center}}\end{minipage}
}
\hspace{2em}
\subfloat[
\textbf{Selection and Inheritance masks} are both necessary.
\textbf{Sharing head} is recommended.
\label{tab:mask_ablation}
]{
\begin{minipage}{0.29\linewidth}{\begin{center}
\tablestyle{1pt}{1.05}
\begin{tabular}{y{80}x{30}x{30}}
Mask type & MAE & MSE  \\
\shline
No mask      &  60.5 & 99.7  \\
$M_i^{S}$ & 675.4 & 809.6 \\
$M_i^{S}$ + $M_i^{I}$  & \baseline{\textbf{54.6}} & \baseline{\textbf{86.9}} \\
\shline
head-dependent &329.2 &396.5\\
head-sharing & \baseline{\textbf{54.6}} & \baseline{\textbf{86.9}} \\
\end{tabular}
\end{center}}\end{minipage}
}
\\
\centering
\caption{\textbf{Ablation experiments} about STEERER on SHHA. If not specified, the default setting for FSIA consists of three components, imposes loss at each resolution, utilizes both inheritance and selection masks, and shares a counting head. Default settings are marked in \colorbox{baselinecolor}{gray}.}

\label{tab:ablations} 
\end{table*}
\begin{figure}[htbp]
	\centering
		\includegraphics[width=0.8\linewidth]{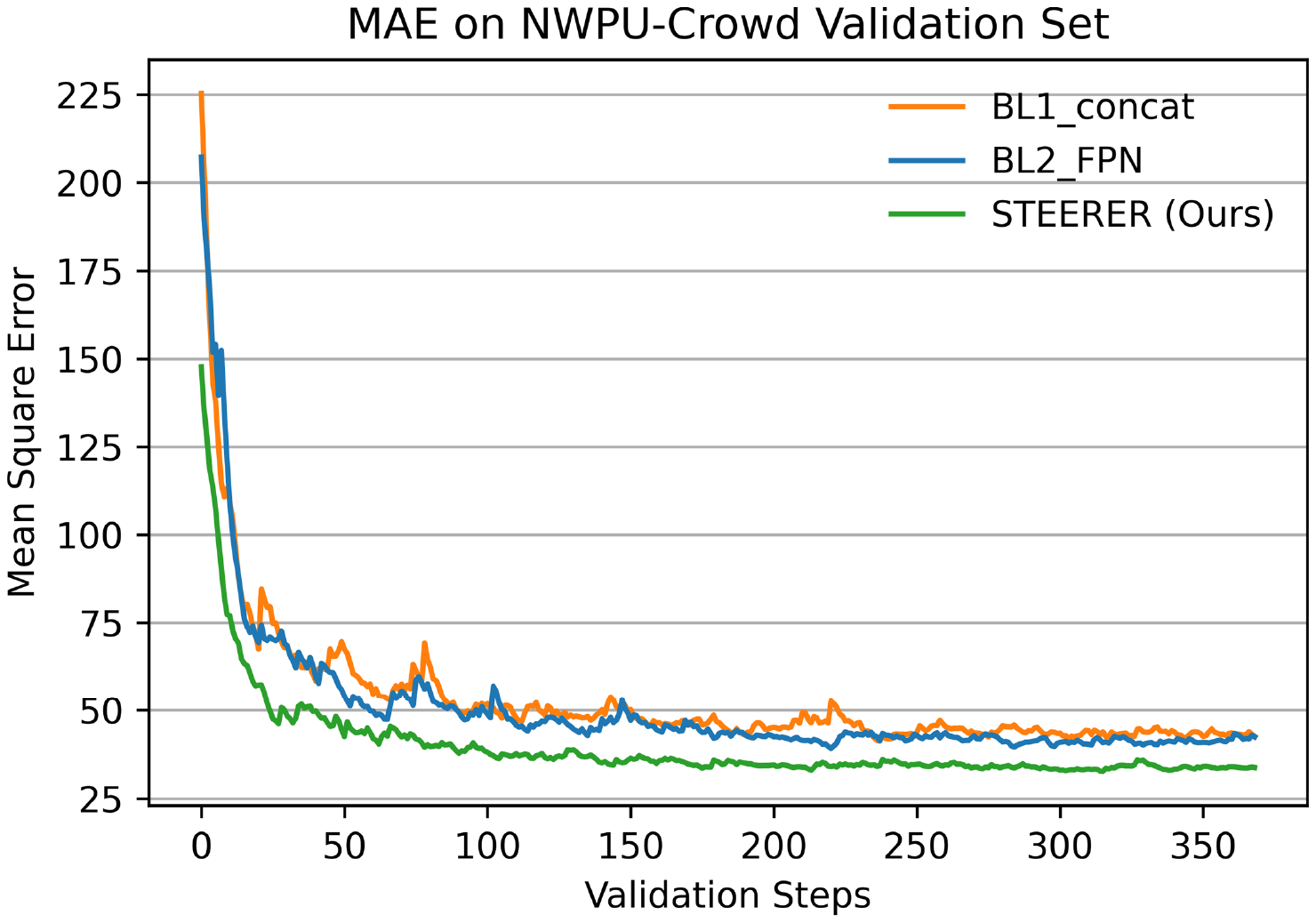}
		\caption{STEERER significantly improves the convergence speed and performance compared with our baselines.}
	\label{fig:coverge}
\end{figure}

\begin{figure}[htbp]
	\centering
		\includegraphics[width=0.98\linewidth]{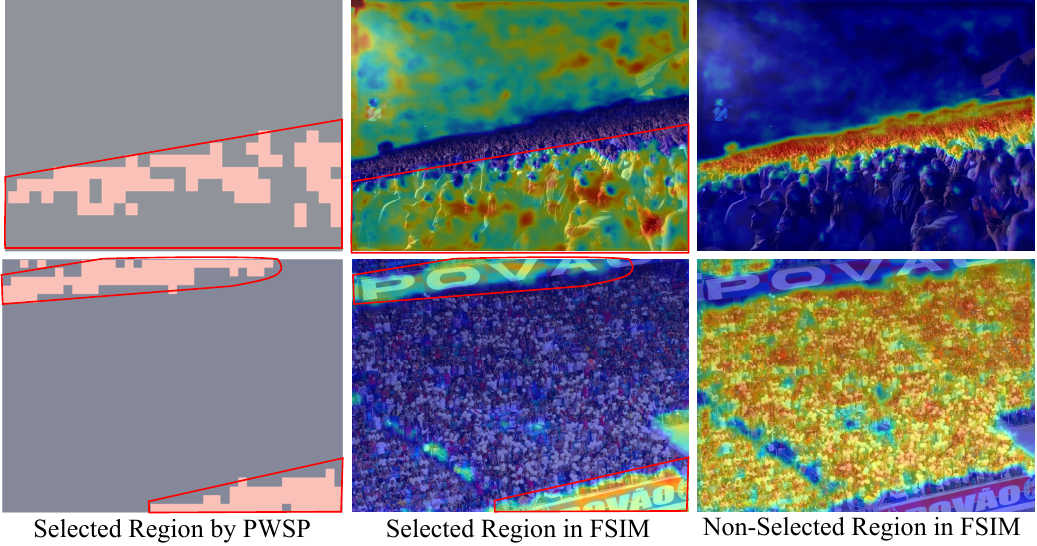}
		\caption{Left masks highlight the region choice by PWSP. Medium figures attentively show FSIA's selected region. The right figures show the complementary scale-uncustomized region.}
		\label{fig:CAM}
\end{figure}

\subsection{Visualization Analysis}
\noindent \textbf{Convergence Comparison.} \cref{fig:coverge} depicts the change  of MAE on NWPU-Crowd  \emph{val set} during training. STEERER and two baselines are trained with the same configuration. This comparison demonstrates that STEERER converges to a lower local minimum with faster speed.    

\noindent \textbf{Attentively Visualize FSIA.} \cref{fig:CAM} shows FSIA's selected results at $R_2$ resolution. The medium Class Activate Map~\cite{zhou2016learning} demonstrates FSIA only selects the large objects in this scale, which aligns with the mask assigned by PWSP. The highlighted regions in the right figures mean objects are small and will be forwarded to a higher resolution.

\noindent  \textbf{Visualization Results.} \cref{fig:vis_all} pictures the quality of the density map and the localization outcomes in practical scenarios. In the crowd scene, both STEERER and the baseline models exhibit superior accuracy in generating density maps in comparison to AutoScale~\cite{Xu_2019_ICCV}. Regarding object localization, STEERER displays greater generalizability in detecting large objects in contrast to the baseline and P2PNet~\cite{song2021rethinking}, as evidenced by the red box in \cref{fig:vis_all}. Additionally, STEERER maintains its efficacy in identifying small and dense objects (see the yellow box in \cref{fig:vis_all}). Remarkably, the proposed STEERER model exhibits transferability across different domains, such as vehicle, tree, and maize localization and counting.

 \subsection{Cross Dataset Evaluation} 
 This cross-dataset testing demonstrates its generalization ability when it comes across scale variations. In \cref{tab:cross_eval}, the baseline model and the proposed model are trained on SHHA (medium scale variations), and then their performance is evaluated on QNRF (large scale variations), and vice versa. Evidently, our method has a higher generalization ability than the FPN~\cite{lin2017feature} fusion method. Surprisingly, our model trained on QNRF with STEERER achieves comparable results with the SoTA results in \cref{tab:counting_results}.

\subsection{Discussions}

\noindent \textbf{Inherit from the lowest resolution.} Fusing from the lowest resolution can hierarchically constrain the density of large objects, while gradually adding the smaller objects. Otherwise, the congested object would be significantly overlapped and challenging to disentangle.

\noindent \textbf{Only output density in the highest resolution.}
STEERER can integrate multiple-resolution prediction maps by leveraging the attention map embedded within FSIM to weigh and aggregate the predictions. However, its counting performance on the SHHA dataset, with an MAE/MSE of 63.4/102.5, falls short of the highest resolution's result of 54.5/86.9.  Also, the highest resolution yields essential details and granularity required for accurate localization in crowded scenes, as exemplified in $\S$ \ref{Sec:counting_exp}.
 
\section{Conclusion} 
In this paper, we propose a selective inheritance learning method to dexterously resolve the scale variations in object counting. Our method, STEERER, first selects the most suitable region autonomously for each resolution with the proposed PWSP and then disentangles the lower resolution feature into scale-customized and scale-uncustomized components by the proposed FSIA in front of the fusion. The scale-customized part is combined with higher resolution to re-estimate the selected region by the last scale and the self-selected region, where those two processes are compacted into selection and inheritance. The experiment results show that our method plays favorably against state-of-the-art approaches with
object counting and localization tasks. Notably, we believe the thoughts behind STEERER can inspire more works to address scale variations in other vision tasks.

\begin{table}
\centering
\caption{Cross dataset evaluation results.}
\resizebox{1.\linewidth}{!}{
\tablestyle{6pt}{0.91}
\begin{tabular}{@{}lccccr}
\toprule 
 Setting & Method & MAE &MSE &NAE \\
\midrule
\multirow{2}{*}{SHHA$\rightarrow$ QNRF} & BL2 & 120.3   &224.5 & 0.1744\\
                                        & STEERER     & \textbf{109.4}   &\textbf{203.4}  & \textbf{0.149}\\

\midrule
\multirow{2}{*}{QNRF$\rightarrow$ SHHA} & BL2 & 58.8  &99.7 & 0.138\\
                                        & STEERER     & \textbf{54.1}   &\textbf{91.2}  & \textbf{0.124}\\
\bottomrule
\end{tabular}
}
\label{tab:cross_eval}

\end{table}

\vspace{15pt}
\noindent \textbf{Acknowledgement}. This work is partially supported by the National Key R\&D Program of China(NO.2022ZD0160100), and in part by Shanghai Committee of Science and Technology (Grant No. 21DZ1100100).

\newpage
{\small
\bibliographystyle{ieee_fullname}
\bibliography{egbib}
}

\end{document}